\documentclass{article}

\usepackage[utf8]{inputenc}
\usepackage[T1]{fontenc}
\usepackage[english]{babel}
\usepackage{microtype}
\usepackage{graphicx}
\usepackage{xcolor}
\usepackage[most]{tcolorbox}
\usepackage{subcaption}
\usepackage{booktabs}
\usepackage{multirow}
\usepackage{siunitx}
\usepackage{natbib}

\usepackage{arxiv}
\usepackage[colorlinks=true,
            linkcolor=black!70!blue,
            citecolor=black!70!blue,
            urlcolor=black!60!blue,
            breaklinks=true]{hyperref}

\renewcommand{\headeright}{Preprint}
\renewcommand{\undertitle}{Preprint}
\renewcommand{\shorttitle}{Can We Trust the Judges?}

\title{Can We Trust the Judges? Validation of Factuality Evaluation Methods
  via Answer Perturbation%
  \thanks{Extended version of Gharsallah et al., \emph{Peut-on faire confiance
  aux juges~? Validation de méthodes d'évaluation de la factualité par
  perturbation des réponses}, Actes de l'atelier EvalLLM 2025, CORIA-TALN 2025,
  Marseille, pp.~228--252. The present version expands the discussion of
  related work; the experimental content is unchanged.}}

\author{
  \bfseries Sarra Gharsallah \quad Adele Robaldo \quad Mariia Tokareva \quad Giovanni Gatti Pinheiro \\
  \bfseries Ilyana Guendouz \quad Raphaël Troncy\thanks{Corresponding author:
    \texttt{raphael.troncy@eurecom.fr}. The work was carried out while all
    authors were at EURECOM; several have since moved to other institutions.}
    \quad Paolo Papotti \quad Pietro Michiardi \\[0.6em]
  \normalfont\normalsize Data Science Department, EURECOM, Campus SophiaTech, Biot, France
}

\date{}

\graphicspath{ {./images/} }

\begin{document}
\maketitle

\begin{abstract}
Evaluating the factual correctness of large language models (LLMs) is vital for many applications. But are our evaluation tools themselves trustworthy? Despite the rise of factuality-based metrics, their sensitivity and reliability remain underexplored. This paper introduces a meta-evaluation framework that systematically tests these metrics using controlled corruptions of gold standard answers. Our method generates ranked outputs with known degrees of degradation to probe how metrics capture nuanced changes in truthfulness. Our experiments reveal that pipeline-based methods, such as the RAGAS's factual correctness metric, better track degradation than LLM-as-judge approaches. We also propose a new variant of the factual correctness metric that provides a competitive and cost-efficient.
\end{abstract}


\section{Introduction}
\label{sec:introduction}
\textbf{Can we trust our tools for evaluating truth?} As large language models (LLMs) become central to retrieving and interacting with information, assessing their correctness is more critical than ever. A growing set of metrics and benchmarks has emerged to meet this need \citep{lin-etal-2022-truthfulqa,min-etal-2023-factscore,ragas2024}, but although we trust these tools to measure truth, we rarely ask how the tools themselves have been validated. In other words, how do we evaluate the evaluators?

LLMs are nowadays deployed in complex, open-ended tasks, such as answering scientific questions and assisting with decision making, where simple multiple-choice answers cannot capture correctness. And yet, many evaluations of LLM performance rely on static \citep{rajpurkar-etal-2016-squad,47761}, test-like benchmarks inherited from earlier NLP tasks. These datasets typically pose closed questions with a single correct answer \citep{hendrycks2021measuring} and ignore the diversity of valid factual completions that LLMs can produce \citep{mirzadeh2025gsmsymbolic,siska-etal-2024-examining}. As a result, there is a growing mismatch between how we evaluate models and how they are actually used.

To address this, researchers have proposed a variety of factuality-based metrics and automatic tools that attempt to quantify the factual correctness of an LLM's output \citep{liu-etal-2023-g,min-etal-2023-factscore}. These include generative LLM-based judges and retrieval-based approaches such as RAGAS. Many of these tools are already widely used to benchmark Retrieval Augmented Generation (RAG) systems, fine-tune models, and power commercial pipelines.

Work on \emph{meta-evaluation} has addressed this question for factual consistency metrics in summarization \citep{gabriel-etal-2021-go,chen-etal-2021-factuality-checkers,honovich-etal-2022-true,ma-etal-2023-bump}, for LLM-based factuality evaluators \citep{chen-etal-2023-felm}, and for evaluator LLMs in general \citep{doddapaneni-etal-2024-finding}. What has not been systematically established is whether the metrics now in routine production use, such as RAGAS-style pipelines and LLM-as-a-judge prompts, track \emph{graded} degrees of factual degradation in open-ended, long-form answers, rather than merely separating consistent from inconsistent text \citep{godbole2025,ramprasad2025}.

Do these factuality-based metrics reliably detect factual errors? Are they sensitive to subtle degradations in truthfulness? Can they distinguish factually accurate completions from incorrect ones in a graded, fine-grained way? For the LLM-based metrics now in production use, the literature has offered little rigorous validation beyond occasional anecdotal case studies or coarse agreement rates \citep{wang-etal-2024-factuality}.

This work addresses this gap by introducing a meta-evaluation framework: A systematic method to test the behavior of factuality-based metrics under controlled conditions. Starting from a gold-standard set of correct answers, we introduce incremental corruptions, such as swapping dates, misnaming entities, and subtly altering claims. This process produces ranked sets of responses with known relationships, which we use to test whether popular factuality-based metrics behave as expected (i.e., rewarding more accurate answers and penalizing less accurate ones).

We make the following contributions:
\begin{itemize}
    \item We propose a controlled corruption pipeline that generates degraded factuality of responses while preserving surface-level fluency and relevance.
    \item We define an evaluation protocol for factuality-based metrics that yields a \emph{totally ordered} set of answers per question, testing sensitivity to degradation, rank correlation, and robustness rather than the binary detection accuracy of prior meta-evaluation benchmarks.
    \item We apply our framework to a suite of popular factuality-based metrics and uncover significant differences in their ability to reflect quality.
\end{itemize}

Our results suggest that some metrics align better with factual degradation, while others are less sensitive to subtle errors. These findings clarify what current tools measure and whether they are up to the task. By exposing the strengths and blind spots of popular metrics, our framework lays the groundwork for more trustworthy LLM evaluation and, ultimately, more reliable systems.

\section{Techniques for Assessing Factual Correctness}
\label{sec:techniques}
In this section, we examine how factuality is currently assessed, reviewing the dominant evaluation approaches, their assumptions, and the extent to which they have been validated.

\subsection{Automatic Factuality Metrics}

Assessing the factuality of large language models (LLMs) is a critical challenge when deploying generative models in high-stakes domains. Traditional automatic metrics, such as BLEU \citep{papineni-etal-2002-bleu} and ROUGE \citep{lin-2004-rouge}, were initially used to assess generated text by measuring the overlap of ngrams with the reference output. However, these metrics do not capture semantic correctness or factual consistency \citep{wang-etal-2020-asking}, and have been shown to correlate poorly with human judgments of truthfulness.

In contrast to learned similarity models, learned similarity metrics such as BERTScore \citep{zhang2020bertscore}, BLEURT \citep{sellam-etal-2020-bleurt}, and COMET \citep{rei-etal-2020-comet} address these limitations leveraging pre-trained or fine-tuned encoders that assess semantic alignment between system and reference outputs. Although these models improve lexical baselines by better capturing fluency and relevance, they often do not distinguish fluent hallucinations from true content, resulting in poor alignment of factuality \citep{chen-eger-2023-menli}.

A more direct approach involves entailment-based metrics, which recast factuality as a Natural Language Inference (NLI) task. FactCC \citep{2020-kryscinski} pioneered this approach by training a classifier on synthetically altered summaries to detect contradictions with the source text. SummaC \citep{laban-etal-2022-summac} applies NLI at sentence granularity, scoring every pair of document and summary sentences. The resulting matrix is then reduced to a single score by averaging, across summary sentences, the highest entailment each one receives. Or one can use a trained aggregator to replace that maximum. Subsequent methods, such as MENLI \citep{chen-eger-2023-menli}, combine NLI-based scores with similarity metrics to resist adversarial perturbations on factuality.

Building on NLI, QA-based methods assess factuality through question-answering. In this setup, a Question Generation (QG) model creates questions from the summary or answer, and a QA model retrieves answers from the source document \citep{wang-etal-2020-asking}. Metrics such as QAGS and QAFactEval \citep{fabbri-2022} compare answer overlaps to detect factual consistency.

Recently, researchers have explored LLMs as evaluators. This technique prompts large models, such as GPT-4o, to directly judge factuality \citep{fu-etal-2024}. These``LLM-as-a-judge'' methods provide flexibility and multi-criteria evaluation but raise concerns around trust, prompt sensitivity, and circularity when the evaluator and generator share architectural biases.

It is also important to review that modular pipeline metrics, such as RAGAS, emerged in the context of Retrieval-Augmented Generation (RAG) systems. Among the many RAGAS metrics, one of our interests is the \textit{factual correctness}. This metric uses LLMs to decompose model answers and a ground truth into atomic claims. Then, it compares each claim for natural language inference via GPT-4o (by default) and computes precision, recall, and an F1 score \citep{ragas2024}, reducing claim-level verdicts rather than averaging entailment scores. These tools support fine-grained error analysis and are tailored for RAG pipelines, but they rely on complex cascades of models whose interactions are not yet fully understood or benchmarked.

The cost of LLM-based verification has also motivated small specialized verifiers: MiniCheck \citep{tang-etal-2024-minicheck} trains a 770M-parameter model on synthetic factual errors and reaches GPT-4-level accuracy on the LLM-AggreFact benchmark at roughly 400 times lower cost. In our work, we investigate a variant of RAGAS factual correctness that shares MiniCheck's cost motivation. In Section~\ref{sec:evaluation}, we introduce a pipeline that uses an off-the-shelf LLM for claim decomposition and a pretrained NLI model for classification; it requires no training data, and either component can be swapped for a newer model as one is released.

\subsection{Validating the Metrics}

Human evaluation remains the gold standard for factuality assessment. Benchmarks such as \textit{SummEval} \citep{fabbri2021}, \textit{FRANK} \citep{pagnoni-2021}, and \textit{TruthfulQA} \citep{lin-etal-2022-truthfulqa} provide expert-annotated judgments of consistency and hallucination, forming the basis for validating automatic metrics. However, while datasets like \textit{TruthfulQA} offer valuable insights into model truthfulness on short, closed-ended questions, they fall short in evaluating more complex, multi-sentence responses grounded in a broader context. For example, \textit{TruthfulQA} primarily targets simple misconceptions or common false beliefs, with a binary framing of correctness.

These annotated benchmarks are also what makes it possible to assess the factuality metrics themselves rather than the systems they score. One thread of that work tests metrics against deliberately corrupted text. For example, GO FIGURE \citep{gabriel-etal-2021-go} formalizes five conditions a factuality metric should satisfy, first among them sensitivity to the degree of factual error. This sensitivity is measured as the Pearson correlation between the number of errors injected into a reference summary and the metric score. Applied to ten metrics, spanning semantic similarity, lexical overlap, QA-based, and cloze-based scoring, the protocol finds ROUGE-1 and ROUGE-L frequently invalid as factuality measures, at times scoring more corrupted summaries as better. AdvFact \citep{chen-etal-2021-factuality-checkers} applies the same idea adversarially, building 24 diagnostic test sets through targeted transformations rather than counting injected errors, and probing six top-performing verifiers with them.

Other benchmarks standardize annotations instead of generating new errors. TRUE \citep{honovich-etal-2022-true} argues that system-level correlation with human judgment is the wrong meta-evaluation target. Instead, it recasts 11 annotated datasets, from summarization to knowledge-grounded dialogue, into a single binary labelling scheme, scored by example-level ROC AUC. It finds that large-scale NLI and QG-QA approaches come out strongest and mutually complementary, well ahead of BARTScore, BERTScore, and n-gram overlap. BUMP \citep{ma-etal-2023-bump} instead contributes 889 human-written minimal pairs with each introducing a single error from an ontology of seven types into a CNN/DailyMail reference summary. Because the pairs are minimal, BUMP can measure \emph{consistency}. It finds that the most discriminative metrics are not necessarily the most consistent. Regarding LLM-generated text specifically, FELM \citep{chen-etal-2023-felm} annotates several hundred ChatGPT responses at segment level across five domains with error types and reference links, reporting that GPT-4-based evaluators detect factual errors far from reliably, with retrieval augmentation helping consistently and chain-of-thought helping only the stronger backbone.

LLM-as-a-judge has also become an object of study \citep{gu-etal-2024-survey}. LLMBar \citep{zeng-etal-2024-llmbar} assembles 419 instruction-following pairs in which one output is objectively preferable, with some written to mislead the evaluator. JudgeBench \citep{tan-etal-2025-judgebench} turns difficult knowledge, reasoning, math and coding datasets into response pairs whose labels follow objective correctness rather than crowdsourced preference, and finds that many strong models, GPT-4o among them, land only slightly above chance. FBI \citep{doddapaneni-etal-2024-finding} instead perturbs answers directly, injecting targeted corruptions across four abilities, factual accuracy among them, to produce 2,400 examples over 22 categories, and finds that LLM judges miss the resulting quality drops in more than half of cases on average, with reference-based evaluation faring better than single-answer or pairwise scoring.

The literature, when taken together, shows that no single factuality metric has yet proven robust, generalizable, and computationally efficient across tasks and domains, least of all in the presence of nuanced errors. Generating reliable synthetic datasets with controlled factual perturbations is therefore fundamental to comparing such metrics. Our work inherits the diagnostic-perturbation methodology of GO FIGURE and AdvFact combined with the meta-evaluation framing of TRUE, differentiating itself along three axes. First, we perturb reference answers in open-domain question answering, where correctness is judged against a gold answer rather than a grounding passage. Prior diagnostic corruptions instead target summarization against a source document. Second, we generate monotonically ordered severity levels supporting rank correlation and exposing whether a metric merely detects errors or actually \emph{orders} them, contrasting with BUMP's minimal pairs, TRUE's binary labels, and FBI's detect-or-miss verdicts. Third, our perturbations are LLM-generated and validated against human experts (Section~\ref{sec:validation}). This makes the protocol inexpensive to re-instantiate on any new gold question-answer set, a practical requirement that hand-built benchmarks such as BUMP cannot meet. The next section presents the pipeline that produces our test datasets. 

\section{The Factual Perturbation Pipeline}
\label{sec:pipeline}
We develop a multi-step \emph{factual perturbation} pipeline. Given a question and its corresponding ground truth answer, it generates alternative answers, ranging from fully correct to progressively incorrect responses. The pipeline produces a series of five answers (A0-A4), where the level A0 corresponds to a faithful paraphrase of the ground truth, and A1 through A4 to increasing levels of factual errors. To illustrate the pipeline capabilities, we present an example extracted from our generated dataset in Figure~\ref{fig:example}. Our approach aims to produce fine-grained levels for evaluating the quality of assessment techniques for natural language responses.

\begin{figure}[h]
 \centering
\begin{tcolorbox}[colback=gray!5!white,colframe=black!75!white,
  title=Example -- \textit{Who did the United States win its independence from?}, fonttitle=\bfseries,
  boxrule=0.8pt, arc=4pt, top=4pt, bottom=4pt, left=6pt, right=6pt]
\footnotesize
\begin{itemize}
    \item \textbf{A0 (Reference)}: Independence Day, commonly known as the Fourth of July or July Fourth, is \textcolor{teal}{a federal holiday} in the United States celebrating \textcolor{teal}{the adoption of the Declaration of Independence} \textcolor{teal}{on July 4, 1776}. \textcolor{teal}{On this day}, the Continental Congress announced that the thirteen American colonies considered themselves a new nation, called \textcolor{teal}{the United States of America}, and were no longer under \textcolor{teal}{British rule}. Interestingly, the Congress had voted to declare independence \textcolor{teal}{two days} earlier, \textcolor{teal}{on July 2}.

    \item \textbf{A1 (Low perturbation)}: Independence Day, commonly known as the Fourth of July or July Fourth, is \textcolor{teal}{a federal holiday} in the United States celebrating \textcolor{teal}{the adoption of the Declaration of Independence} \textcolor{red}{on August 5, 1776}. \textcolor{teal}{On this day}, the Continental Congress announced that the thirteen American colonies considered themselves a new nation, called \textcolor{teal}{the United States of America}, and were no longer under \textcolor{teal}{British rule}. Interestingly, the Congress had voted to declare independence \textcolor{teal}{two days} earlier, \textcolor{teal}{on July 2}.

    \item \textbf{A2 (Medium perturbation)}: Independence Day, commonly known as the Fourth of July or July Fourth, is \textcolor{teal}{a federal holiday} in the United States celebrating \textcolor{teal}{the adoption of the Declaration of Independence} on \textcolor{red}{August 5, 1781}. \textcolor{red}{On that moment}, the Continental Congress announced that the thirteen American colonies considered themselves a new nation, called the \textcolor{teal}{United States of America}, and were no longer under \textcolor{teal}{British rule}. Interestingly, the Congress had voted to declare independence \textcolor{red}{several days} earlier, \textcolor{teal}{on July 2}.

    \item \textbf{A3 (High perturbation)}: Independence Day, commonly known as the Fourth of July or July Fourth, is an \textcolor{red}{unofficial event} in the United States celebrating \textcolor{red}{a proposal of} \textcolor{teal}{the Declaration of Independence} \textcolor{red}{on August 5, 1781}. On that moment, the Continental Congress announced that the thirteen American colonies considered themselves a new nation, called the \textcolor{teal}{United States of America}, and were no longer under \textcolor{teal}{British rule}. Interestingly, the Congress had voted to declare independence \textcolor{red}{several days} earlier, \textcolor{red}{on August 2}.

    \item \textbf{A4 (Extreme perturbation)}: Independence Day, commonly known as the Fourth of July or July Fourth, is \textcolor{red}{an unofficial event} in the United States celebrating \textcolor{red}{a proposal of the drafting of Independence} on \textcolor{red}{August 5, 1781}. \textcolor{red}{On that moment}, the Continental Congress announced that the thirteen American colonies considered themselves a new nation, called \textcolor{red}{the United States of the Colonies}, and were no longer under \textcolor{red}{Spanish rule}. Interestingly, the Congress had voted to declare independence \textcolor{red}{several days} earlier, \textcolor{red}{on August 2}.
\end{itemize}
\end{tcolorbox}
 \caption{Example of a paraphrased reference answer and four variants introducing an increasing level of factual errors.}
 \label{fig:example}
\end{figure}

The pipeline operates as follows:
\begin{itemize}
    \item \textbf{Paraphrasing Ground Truth (A0 Generation):} The pipeline uses an LLM to paraphrase the ground truth while ensuring semantic equivalence. This step ensures linguistic diversity without altering correctness.
    \item \textbf{Factual Component Extraction:} The pipeline uses syntactic and dependency parsing of A0 to identify the most significant factual components -- such as entities, quantities, and modifiers -- that contribute to the meaning of the answer. These components are tagged within the answer text.
    \item \textbf{Question-Answer Term Filtering:} To avoid the answers from drifting from the original question, the pipeline filters factual components overlapping with the question contents, ensuring that modifications target information not explicitly recoverable from the question alone.
    \item \textbf{Importance Ranking:} The pipeline uses an LLM to rank the remaining factual components by importance order in the answer context. Ranking emphasizes components that define the main event, actors, implications, or numerical details, while de-emphasizing vague or ancillary information. From our experience, powerful LLMs do a fair job in categorizing the major points.
    \item \textbf{Component Selection:} The top-ranked factual components are retained based on a tunable threshold (e.g., top 80\%). These serve as candidates for subsequent modification.
    \item \textbf{Grouping Factual Elements:} The remaining factual components are grouped into semantically coherent sets, each mapped to a distinct ``factual perturbation levels'' (A1--A4), such that each set corresponds to a degree of factual deviation. Each group balances high- and low-importance components to ensure that each perturbation level reflects a controlled and progressively more severe deviation. The goal of this step is to spread the degree changes more or less evenly between the four levels.
    \item \textbf{Controlled Perturbation and Answer Generation (A1–A4):} Starting from $A0$, the pipeline incrementally selects grouped factual elements and requests an LLM to modify the factual items in a plausible but subtly incorrect way.
\end{itemize}

Corrupting a reference text as an evaluation instrument has precedent. For example, GO FIGURE \citep{gabriel-etal-2021-go} injects entity, pronoun, and verb-negation errors using part-of-speech tagging and named entity recognition, together with WordNet-based antonym substitution of adjectives. Or, AdvFact \citep{chen-etal-2021-factuality-checkers} applies adversarial transformations for the same purpose. Our pipeline differs in delegating the substitution step to an LLM under a plausibility constraint, allowing injected errors to remain fluent and contextually credible. It also partitions perturbations into ordered severity groups, rather than the unordered error counts GO FIGURE and AdvFact report.

Appendix \ref{appendix:pipeline-details} provides more detailed explanations about each step; our code and datasets are publicly available\footnote{\url{https://github.com/GiovanniGatti/truthbench}, archived at \url{https://doi.org/10.5281/zenodo.22299251}\label{footnote:repo}}. This structured perturbation framework enables fine-grained benchmarking of factuality evaluation methods under controlled degradation scenarios.

\section{Validating the Pipeline}
\label{sec:validation}
As described in the previous section, our pipeline employs LLMs for certain tasks. These LLMs can themselves produce unexpected and inaccurate outputs, which could degrade the quality of the factual perturbation. In fact, BUMP \citep{ma-etal-2023-bump} relies entirely on human-authored minimal edits precisely because model-generated errors were found to be systematically different from human-generated ones. For this reason, we validate the pipeline by testing whether the LLM-generated perturbations are distinguishable in quality from expert-written ones at each severity level.

One way to do so is to request human experts to perform the same task: From a gold Q\&A dataset (i.e., questions and ground truths), produce several variants (A0--A4) with increasing degree of incorrectness. Then, we can request a second group of experts to compare the quality of those outputs with those generated by the factual perturbation pipeline.

We requested two evaluators to compare the quality of the factual perturbation introduced in a blind-randomized test. For each question, the evaluators see the five levels of answers. They can also see the human-expert and pipeline-generated versions for each level side by side. The evaluators are blind to the source (pipeline or expert) of each answer and are asked to choose which one best aligns with the intended perturbation level. They can also opt for accepting or rejecting both alternatives. To facilitate the evaluation procedure, we provide evaluators with a user-friendly interface (UI) that includes scoring guidelines and quick visualization of the differences between perturbation levels. More information about the UI is available in the Appendix \ref{appendix:evaluation-interface}.

The evaluators accessed 20 Q\&A pairs (a total of 100 A0--A4 pairs generated by a human expert and the factual perturbation pipeline). The contingency table of their assessment choices is available in Table \ref{tab:contingency-table}.

\begin{table}[htbp]
\centering
\renewcommand{\arraystretch}{1.15}
\begin{tabular}{lcccc}
\toprule
\textbf{Evaluators} & \textbf{AI} & \textbf{Both are good} & \textbf{Both are bad} & \textbf{Expert} \\
\midrule
\textbf{AI}              & 6  & 14 & 0 & 0  \\
\textbf{Both are bad}    & 0  & 2  & 0 & 1  \\
\textbf{Both are good}   & 7  & 32 & 0 & 5  \\
\textbf{Expert}          & 2  & 8  & 0 & 3  \\
\bottomrule
\end{tabular}
\caption{Contingency table of human preferences (excluding A0). Rows indicate assessments by one evaluator, and columns by the other.}
\label{tab:contingency-table}
\end{table}

\begin{figure}[t]
\vspace{8pt}
\centering
\begin{subfigure}[t]{0.55\linewidth}
    \centering
    \def\svgwidth{\linewidth}
    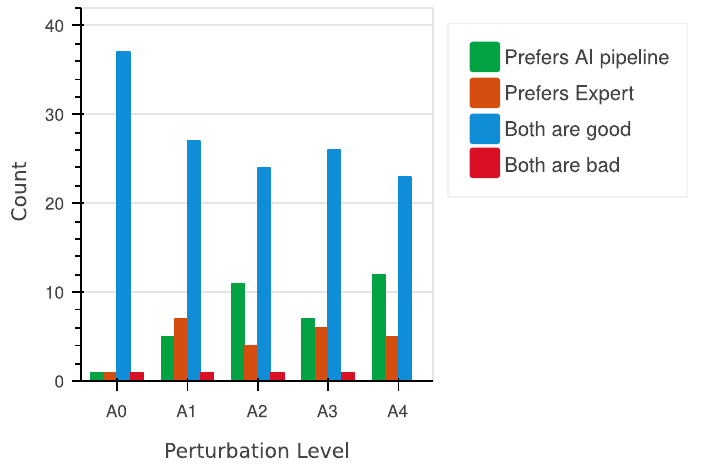 
    \caption{Expert assessment breakdown.}
    \label{fig:by-level}
\end{subfigure}
\begin{subfigure}[t]{0.42\linewidth}
    \centering
    \def\svgwidth{\linewidth}
    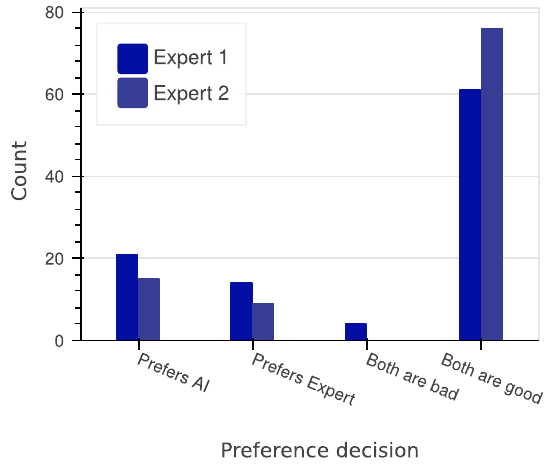 
    \caption{Individual expert assessments.}
    \label{fig:by-expert-preference}
\end{subfigure}
\caption{Evaluation of AI pipeline vs. human expert introduced factual perturbation in a blind-randomized test.}
\end{figure}

We focus on evaluating the non-inferiority of the AI pipeline with respect to human experts. Most pairs (85\%) were rated as ties, i.e., there is no clear distinction in quality between the AI pipeline and expert-generated answers. In fact, the level A0 (see Figures \ref{fig:by-level}) shows a nearly universal equivalence rate (95\% ties), so we exclude it from further analysis.

For the remaining factual perturbation levels ($N = 80$ pairs), we employ a composite scoring system ($-1$: Expert preference, $+1$: AI preference, $0$: ties/mixed preferences) and predefined a non-inferiority margin of $\delta = -0.1$. This margin reflects a conservative threshold, where AI performance would not significantly lag behind human performance in real-world applications.

The AI pipeline achieves a mean composite score of $0.125$ (90\% bootstrap CI: $[0.025, \infty]$), with the lower confidence bound exceeding the non-inferiority margin. This indicates statistical support for non-inferiority. Notably, 82.5\% of evaluations resulted in ties (e.g., both evaluators selecting ``Both are good'' or one expressing uncertainty, see Figure \ref{fig:by-expert-preference}), suggesting frequent perceptual equivalence between AI and human responses. Contingency tables reveal minimal direct disagreement: Only $2$ of the $80$ examples ($2.5\%$) showed opposing preferences (AI vs. Expert). However, inter-rater reliability remained low (Cohen’s $\kappa = 0.113$), reflecting challenges in consistently distinguishing between AI and human outputs. Low inter-rater agreement on this kind of task is not unusual: TRUE \citep{honovich-etal-2022-true} found annotation errors in 35 of the 80 examples that all three of its best-performing metrics had misclassified, indicating that expert judgment of factual consistency is itself unreliable on the hardest cases.

Post hoc power analysis estimates $77\%$ power to detect non-inferiority at $\alpha = 0.05$, suggesting moderate sensitivity. While sufficient to support our hypothesis, larger samples ($\sim 50$ examples) would strengthen reliability.

Despite this borderline statistical significance, the AI pipeline demonstrates non-inferiority to human experts during a blind evaluation under perturbed conditions (A1–A4), with high equivalence rates and statistical bounds supporting its functional interchangeability in most cases.

\section{Evaluation of LLM Assessment Methods}
\label{sec:evaluation}
We leverage our pipeline to gain deeper insight into how the techniques presented in Section \ref{sec:techniques} behave regarding factual correctness. We apply it to the gold dataset to generate progressively factually perturbed alternatives. These perturbed alternatives are then fed into RAGAS and LLM-as-judge. By comparing the final factuality scores with the intended perturbation level, we can evaluate how well each technique performs in detecting factual errors in open-ended answers.

We expect that the factuality scores decrease as the perturbation increases. Thus, we evaluate performance using two metrics: \textbf{Pearson correlation}, which measures the linear relationship between perturbation levels (A0 to A4) and factuality scores, and \textbf{Kendall's tau}, which assesses whether the relative ranking of factuality scores correctly reflects the increasing perturbation levels. We hypothesize that models with stronger negative correlation (closer to $-1$) better capture factual degradation.

We generate 500 A0 to A4 examples using 100 Q\&A from the Google Natural Questions dataset \citep{47761} (i.e., the gold dataset). This dataset consists of over 300,000 Q\&A queries sourced from Google search, with long answers typically drawn from Wikipedia. To ensure high-quality examples for evaluation, we sample the first 100 complete questions (i.e., ending with a period) whose long answer is longer than 250 characters and shorter than 700 characters.

We evaluate six state-of-the-art, open-weights LLMs of various sizes and providers. In addition, we assess the factual correctness pipeline described in Section \ref{sec:techniques}, using both the default RAGAS implementation based on GPT-4o-mini \citep{ragas2024}, and a variant that uses an open-weight LLM for sentence splitting and a Natural Language Inference (NLI) model (\texttt{nli-deberta-v3-large} \citep{reimers-2019-sentence-bert}) for sentence classification (LLM + NLI). This diverse setup allows us to explore a wide range of models and techniques, comparing their performance in terms of both computational efficiency and accuracy in detecting factual errors. Moreover, we use a reference-based LLM-as-a-judge protocol (Appendix~\ref{appendix:prompts}), the configuration FBI \citep{doddapaneni-etal-2024-finding} found most reliable among judge protocols.

\begin{table}[t]
\centering
\renewcommand{\arraystretch}{1.15}
\begin{tabular}{llccc}
\toprule
\textbf{Method} & \textbf{LLM} & \textbf{Pearson (95\% CI)} & \textbf{Kendall} & \textbf{Kendall (95\% CI)} \\
\midrule
\multirow{6}{*}{LLM-as-judge} 
  & gemma3: 4b & -0.63 [-0.69, -0.58] & -0.79 & [-0.82, -0.77] \\
  & llama3.3: 70b & -0.74 [-0.78, -0.70] & -0.86 & [-0.88, -0.84] \\
  & mistral-small3.1: 24b & -0.71 [-0.75, -0.66] & -0.76 & [-0.79, -0.72] \\
  & phi4: 14b & -0.74 [-0.78, -0.70] & -0.81 & [-0.83, -0.78] \\
  & prometheus-v2: 7b & -0.62 [-0.67, -0.56] & -0.70 & [-0.75, -0.66] \\
  & qwen2.5: 7b & -0.63 [-0.68, -0.57] & -0.72 & [-0.76, -0.67] \\
\midrule
RAGAS  
  & gpt-4o-mini & \textbf{-0.87} [-0.90, -0.85] & -0.95 & [-0.97, -0.93] \\
\midrule
\multirow{2}{*}{LLM + NLI} 
  & gemma3: 12b & -0.82 [-0.85, -0.79] & \textbf{-0.96} & [-0.98, -0.94] \\
  & llama3.3: 70b & -0.83 [-0.86, -0.80] & -0.94 & [-0.96, -0.92] \\
\bottomrule
\end{tabular}
\caption{Correlation between factual perturbation levels and factuality score.}
\label{tab:correlation-results}
\end{table}

Table \ref{tab:correlation-results} presents the correlation coefficients for each technique, quantifying how well their factuality scores track the perturbation levels. From the table, we can observe several important trends that shed light on the relative performance of each technique.

First, methods using LLM-as-a-judge generally exhibit weaker performance compared to the other techniques. These models consistently show lower Pearson correlation and Kendall’s tau values, indicating that they are less effective at capturing the relationship between perturbation levels and factuality scores. This suggests that LLM-as-a-judge methods are not as reliable in detecting factual errors introduced by increasing perturbation, which may be due to their reliance on generative models rather than more explicit approaches.

The methods using an explicit pipeline for calculating factual correctness demonstrate much stronger performance, both in terms of Pearson’s correlation and Kendall’s tau. These models achieve significantly stronger Pearson correlation values, indicating a stronger linear relationship between increasing perturbation levels and the corresponding factuality scores. Furthermore, the strong Kendall’s tau highlights the ability of these methods to maintain the correct ranking of factuality across different perturbation levels. This suggests that these methods are better at detecting and maintaining consistency in factuality, even as perturbation is added. The strong Pearson and Kendall’s tau values together indicate a more reliable and consistent assessment of factual correctness.

Moreover, the use of open-weights LLMs for sentence splitting combined with a dedicated NLI model (LLM + NLI) offers an attractive alternative to proprietary models based on \texttt{gpt-4o-mini}. Despite not always outperforming in terms of Pearson correlation, these models provide a more computationally efficient and cost-effective solution while maintaining strong performance in detecting factual errors.

\begin{figure}[htbp]
\centering
\begin{subfigure}[t]{0.48\linewidth}
    \centering
    \def\svgwidth{\linewidth}
    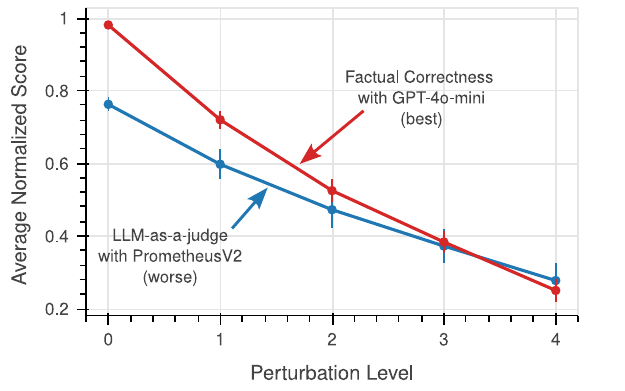
    \caption{Average score vs. perturbation level}
    \label{fig:score-vs-noise}
\end{subfigure}
\begin{subfigure}[t]{0.48\linewidth}
    \centering
    \def\svgwidth{\linewidth}
    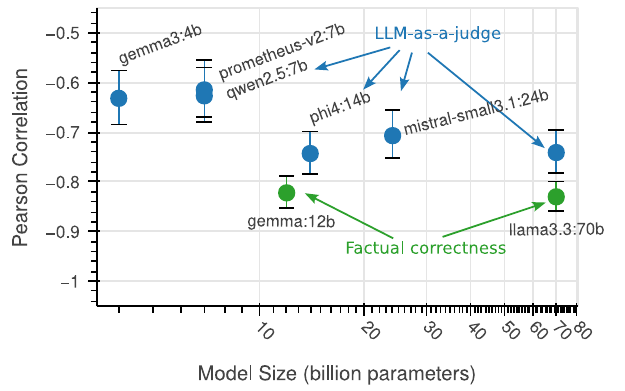
    \caption{Pearson correlation vs. LLM size}
    \label{fig:model-size-vs-pearson-correlation}
\end{subfigure}
\caption{Performance of factuality evaluation methods.}
\end{figure}

Figure \ref{fig:score-vs-noise} provides another perspective on how the assessment techniques distinguish answers across perturbation levels. The red curve (GPT-4o-mini with RAGAS) steadily declines from 1 to approximately 0.2 as perturbation increases, closely mirroring the structure of the perturbation generation process, where 20\% of the facts are preserved. This confirms that the factual correctness pipeline not only correlates well with ground truth but also faithfully tracks factual degradation, utilizing the full score range in a way that is interpretable and grounded. In contrast, PrometheusV2 \citep{kim-etal-2024-prometheus} used as a judge (blue curve) shows a flattened response, often over penalizing correct answers or failing to penalize factual errors adequately. Although explicitly fine-tuned for comparing responses to score rubrics, its output as a factuality scorer does not reflect systematic changes in data quality, undermining its usefulness for fine-grained evaluation. This observation underscores that factual correctness methods, like RAGAS, not only achieve better alignment with the intended ranking (as shown by Kendall's tau) but also produce score distributions that are more calibrated to the factual quality of responses. These insights reinforce the overall advantage of factual correctness pipelines for this task.

Finally, in terms of model size, Figure \ref{fig:model-size-vs-pearson-correlation} reveals that larger models do not necessarily lead to better evaluation performance. In fact, several mid-sized open-weights models achieve higher correlations than much larger judge-based LLMs. This indicates that accurate factuality evaluation does not strictly require massive LLMs, and that smaller, targeted models within structured pipelines may offer both better performance and higher efficiency. Furthermore, the LLM size also seems to have almost no impact on the final performance of the factual correctness evaluation method. These findings reinforce that thoughtful pipeline design and model selection can outperform brute-force scaling, offering a more practical and robust approach to the evaluation of factuality. However, our findings do not contradict that scaling helps \emph{within} a fixed metric design. In fact, TRUE \citep{honovich-etal-2022-true} reports consistent gains from larger backbones across its NLI, BLEURT and BERTScore variants. Our results instead indicates that the choice of evaluation architecture dominates backbone scale across designs.

\section{Known Limitations}
\label{sec:limitations}
Our pipeline systematically applies linguistic and semantic modifications using dependency parsers and predefined operators. However, the effectiveness of these perturbations can vary depending on the properties of the target text. For instance, verbose or highly detailed answers, such as those generated by LLMs, may require more targeted or intensive perturbations to produce noticeable semantic shifts, whereas shorter, more concise answers may be more sensitive to minor changes. As a result, ``uniformity'' in the degree of perturbation across questions and ground truths is not guaranteed.

Furthermore, specific perturbations may not always affect the core content of the answer. Changes may preserve the underlying meaning despite surface-level changes (see "Who breaks a tie in the US Senate?" example in Appendix \ref{appendix:dataset-examples}). While we do not have quantitative evidence that such cases are prevalent across the dataset, this example highlights a possible limitation of perturbing verbose answers where the core fact is only a small part of the text. In fact, the two evaluators had no specific instructions on whether they should accept or reject such cases. Further analysis is warranted to determine how often such cases occur and whether alternative perturbation strategies are needed. Conversely, some perturbations may introduce inconsistencies or semantic contradictions within the answer (such as in "Who wrote the text for \textit{Jeanie with the Light Brown Hair}?" example in Appendix \ref{appendix:dataset-examples}). For such examples, they fail the semantic guidelines and should be counted as a rejected example (i.e., either accept the human alternative or reject both).

Our work measures sensitivity to genuine factual degradation, but we do not test robustness to fact-preserving manipulation. Specialized entailment classifiers such as AlignScore and MiniCheck can be inflated by appending content-free sentences (e.g., ``the document discusses''), whereas LLM-based prompting is comparatively robust \citep{ramprasad2025}. The two findings are compatible: a metric can track degradation faithfully while remaining exploitable by inputs that do not degrade content, and a full characterization of an evaluation metric requires both tests. Extending our pipeline with fact-preserving perturbations as a control condition is a natural next step.

Finally, while our method is language-agnostic in principle, it depends on the availability of reliable dependency parsers and LLMs for the target language. Languages with complex morphology or syntax, or with low resources, may suffer from perturbation accuracy and coverage. We have not yet thoroughly evaluated the reasoning capability of recent LLMs triggered by specific prompt instruction (e.g. \texttt{<think>}) and chain-of-thoughts decomposition that could also improve the factuality assessment.

\section{Conclusions}
Our experiments reveal that pipeline-based methods for factual correctness, such as RAGAS, significantly outperform LLM-as-judge approaches in detecting factual errors. These methods exhibit stronger Pearson correlation and Kendall's tau values, indicating their superior ability to track factual degradation and maintain consistency in ranking as perturbation increases. We also find that open-source LLMs combined with natural language inference (NLI) models offer a cost-effective alternative. Although these models may not consistently outperform the larger LLMs in Pearson's correlation, they still demonstrate competitive performance in detecting factual errors, offering a more computationally efficient solution. In terms of model size, larger LLMs do not necessarily lead to better evaluation performance. In fact, several mid-sized open-source models achieve higher correlations than much larger judge-based LLMs. This highlights the importance of thoughtful pipeline design and model selection over simply scaling the model size. Furthermore, our findings suggest that structured, smaller models within effective pipelines can outperform larger models in factuality evaluation.

Beyond empirical findings, our work proposes a general-purpose open-source pipeline for benchmarking factuality metrics. Given a reference dataset, practitioners can use our framework to test and compare multiple evaluation metrics and select the one that best aligns with the specific requirements of their task. Furthermore, our approach helps standardize the comparison of metrics across the community, raising the standards of trust and rigor in assessment research. The pipeline is publicly available for the community to use\footref{footnote:repo}.

\section*{Acknowledgments}
This work has been partially supported by the French Public Investment Bank (Bpifrance) within the LettRAGraph project. It also supports our broader effort in developing a Socratic companion assisting students when learning educational resources~\citep{eurecom-aixedu2024}.

%%================================================================
%% Note : si l'on préfère éviter de factoriser les crossrefs :
%% bibtex -min-crossrefs 99 taln-exemple
%%================================================================
\bibliographystyle{plainnat}
\bibliography{biblio}

\newpage

\appendix

\section{Details on the Factual Perturbation Pipeline} \label{appendix:pipeline-details}

This section describes each step of the processing pipeline used in the generation of paraphrased answers and their perturbed variants. Each step performs a targeted transformation or filtering operation, contributing to the structured manipulation of factual information for evaluation purposes.

\subsection*{Paraphrasing}
The initial step rewrites the reference answer to introduce lexical and syntactic diversity without altering its factual content. The goal is to preserve all informational elements while presenting them in different linguistic forms. This allows subsequent stages to operate on expressions that are not verbatim replications of the original data, thereby increasing the realism of downstream variations. This step is performed with the assistance of an LLM which prompt is available in Appendix \ref{appendix:prompts}.

\subsection*{Factual Span Identification}
This step identifies a set of textual spans within the paraphrased answer that are likely to carry factual content. The process involves parsing the sentence structure and extracting spans based on their syntactic role and lexical category. Specifically, the method targets constituents such as:

\begin{itemize}
    \item Direct objects, attributes, and complements of the main verb (e.g., "the new policy" in "The government announced the new policy.");
    \item Adverbial modifiers and prepositional phrases that express circumstantial detail (e.g., "in 2021", "with high confidence");
    \item Appositional phrases and descriptive noun modifiers;
    \item Numerical expressions that are not part of the sentence subject.
\end{itemize}

Spans that fall under the grammatical subject are excluded to avoid catastrophic modifications (e.g., ``Maria left the room.'' becomes ``Bob left the room'') . In cases where a noun phrase subject contains embedded relative clauses (e.g., ``the researchers who conducted the trial''), the pipeline uses only the head of the subject (e.g., ``the researchers'') to determine exclusion, and modifiers in the relative clause (e.g., ``the trial'') remain eligible.

Candidate spans are constructed by identifying tokens that fulfill one of several dependency-based heuristics and then expanding to cover the minimal complete phrase headed by that token. This yields noun phrases, quantified expressions, or adverbial constructions that are syntactically cohesive and semantically atomic. Coordination is handled by propagating eligibility from a conjunct to its siblings when the head of the coordination satisfies one of the criteria. Nested or overlapping spans are suppressed to ensure a non-redundant set of annotations.

The resulting spans are bracketed in the text, forming a structured intermediate representation used in subsequent steps for filtering, ranking, and perturbation.

\subsection*{Question-Based Filtering}

To prevent the pipeline from altering content that is central to the meaning of the original question, this step removes any factual span that shares lexical items with the question itself. The rationale is that modifying such content may directly interfere with the core referents or intent of the question, leading to a catastrophic misleading outputs. For instance, in a question like ``Why is the sky blue?'', altering spans containing the word ``blue'' in the corresponding answer (e.g., ``The sky appears blue because'') could result in a breakdown of semantic coherence.

The filtering process proceeds as follows:

\begin{enumerate}
    \item \textbf{Tokenization and Normalization:} The question text is tokenized using a simple heuristic tokenizer that splits on whitespace and removes punctuation. All tokens are lowercased to ensure case-insensitive comparison.
    \item \textbf{Stop Word Removal:} Common English stop words (e.g., the, is, and, why) are removed from the tokenized question. This step helps to focus the comparison on semantically meaningful words that are more likely to carry content, rather than grammatical function.
    \item \textbf{Span Comparison:} Each factual span identified in the previous step is compared to the filtered set of question words. If any non-stopword token from a span appears in the filtered question word list, the span is excluded from further processing. This conservative strategy ensures that even partial lexical overlap — e.g., a span like ``sky temperature'' in a question containing ``sky'' — triggers exclusion.
\end{enumerate}

The result is a subset of factual spans that are disjoint, in terms of lexical content, from the question. This helps preserve the fidelity of the modified answers by shielding question-relevant information from unintended alteration.

\subsection*{Factual Relevance Ranking}

To prioritize factual content by importance, each identified span is ranked according to its semantic contribution to the sentence. The goal is to surface spans that convey what the sentence is about, who is involved, what happens, and any concrete facts (e.g., quantities, dates, consequences), while downranking vague or auxiliary spans.

The input to this step is the paraphrased sentence with factual spans enclosed in brackets and annotated with numeric indices (e.g., [term:0], [term:1], ...). A language model is prompted with this version of the sentence and asked to return an ordered list of indices, ranked by factual relevance. The prompt (available in Appendix \ref{appendix:prompts}) includes an in-context example and invites the model to reflect on its ranking strategy within <thinking></thinking> tags, followed by a final list within <output></output>. Even though we do not use ``thinking models'', we found that the thinking tag significantly improved reliability of outputs.  

The pipeline parses and validates the output, ensuring it is a complete permutation of the input indices. On failure, the query is retried a fixed number of times. The resulting ranking is then used to structure subsequent filtering and perturbation steps, allowing the pipeline to focus edits on less critical content.

\subsection*{Selective Retention of Factual Content}

Following the relevance ranking of factual spans, the pipeline retains only the top portion of these spans for downstream modification. This step assumes that the highest-ranked spans encode the most essential information, while lower-ranked spans are more safely altered or removed without compromising the core meaning.

Retention is controlled by a parameter $k \in (0,1]$ that determines the proportion of top-ranked spans to keep. Given a ranked list of $n$ spans, the pipeline selects the first $\left\lceil k \cdot n \right\rceil$. These retained spans are treated as semantically central and preserved in all subsequent perturbations. This filtering acts as a preventive guardrail, ensuring that less critical elements (often connective elements) are not unnecessarily modified.

\subsection*{Grouping Factual Elements}
To perform structured perturbations, the retained factual spans are grouped into subsets that will be used to generate controlled variants of the answer. The goal of this step is to ensure changes between A0-A4 are gradual (i.e., avoiding perceived big jumps between one level and another).

The grouping procedure follows a balanced batching strategy. The pipeline first creates a sequence of span indices ordered by descending importance. This sequence is then divided into $l$ groups (i.e., the number of variants to be generated) using a zigzag batching pattern. Specifically:

\begin{enumerate} 
\item The list of indices is split into chunks of size $l$, producing an initial batch matrix. 
\item These batches are then interleaved into $l$ groups such that each group receives a different element from each batch. 
\item To mitigate ordering bias, the final groups are randomly shuffled before use.
\end{enumerate}

This method ensures that each group contains a mixture of more and less critical spans, enabling the generation of outputs that vary in both content and degree of perturbation. Moreover, by enforcing disjoint sets across groups, the pipeline avoids redundant modifications and ensures that each variant captures a unique transformation trajectory.

The resulting groups serve as the basis for the controlled perturbation in the subsequent step.

\subsection*{Generating Controlled Perturbation}

The final stage of the pipeline introduces controlled perturbations to the paraphrased answer by selectively modifying factual spans retained and grouped in the previous step. The objective is to produce minimally edited variants that simulate realistic factual errors while preserving surface-level fluency and grammatical correctness.

This step operates in multiple rounds, each corresponding to a different perturbation level. At each level, the pipeline selects a distinct subset of the factual spans. The items selected for mofification are marked between the square brackets (e.g, [ ]). The other factual items are marked with double-bracket notation (e.g., {{term}}). The choice of double-brackets turned out to be important, as other character sequences (such as angle brackets < and other symbols used in HTML and XML) delivered inconsistent results. The final text is then passed to an LLM, accompanied by a prompt that instructs the model to replace the terms between square brackets with plausible but incorrect or misleading alternatives. The prompt is designed to encourage the LLM to create alternatives for the edits (using <thinking> tags) and then produce a fully edited output (enclosed in <output> tags). See Appendix \ref{appendix:prompts} for the full prompt template.

The key properties of the generation of perturbation process include:

\begin{itemize} 
\item \textbf{Iterative refinement:} Each perturbation level builds upon the previous one. That is, the output of level $i$ becomes the input to level $i+1$, allowing for compounding perturbations that remain locally coherent. 
\item \textbf{Groupwise perturbation:} The set of non-retained spans is partitioned into mutually exclusive groups using a zig-zag round-robin strategy. Each level modifies only one group, ensuring diversity of edits while avoiding over-concentration of perturbation in any single region of the text. \item \textbf{Reinsertion and postprocessing:} The LLM’s output is parsed to extract the altered sentence. Any remaining masked tokens are converted back into bracketed form (i.e., [term]) for internal consistency, and the cleaned version is saved as a finalized perturbed variant. 
\end{itemize}

The result is a sequence of perturbed answers A1 to A4, each differing from the original paraphrase by an increasing number of factually altered elements. These variants are suitable for use in evaluation tasks such as robustness testing, factuality classification, or error localization.

\newpage
\section{System prompts} \label{appendix:prompts}

\begin{tcolorbox}[title={LLM paraphrasing}, colback=blue!10]
Rewrite the provided sentence to express the same idea in slightly different words while preserving full accuracy, completeness, and meaning. Ensure the content remains faithful to the original and includes all key details. Do not add any note.

Original:
\{ground\_truth\}

Paraphrased version:
\end{tcolorbox}

\begin{tcolorbox}[title={LLM ranking}, colback=blue!10]
Output the indexes of terms in square brackets [ ] from the text between triple backticks \textasciigrave\textasciigrave\textasciigrave by terms that shape what the text is about, who it involves, consequences, hard numbers, dates, and facts. Downrank marked terms that are vague references, general connectors, or dependent on other terms in square brackets. You are given a free space to decide your ranking strategy between the tags <thinking></thinking>\\

Example:\\

\textasciigrave\textasciigrave\textasciigrave

Recent studies have shown [a correlation:0] between [social media use:1] and [increased anxiety:2] among [teenagers:3]. Although some researchers argue that online interaction can promote [social connection:4], others warn about [its impact:5] on [self-esteem:6] and [sleep patterns:7]. Debates intensified after a [whistleblower:8] revealed internal data from [a major tech company:9] indicating [awareness:10] of [these risks:11].

\textasciigrave\textasciigrave\textasciigrave\\

<thinking>

The main terms are [social media use:1], [teenagers:3], [a major tech company:9]...Did I forget any missing terms?...

</thinking>\\

OUTPUT: [1, 3, 9, 2, 8, 6, 7, 4, 10, 0, 11, 5]
\end{tcolorbox}

\newpage

\begin{tcolorbox}[title={LLM perturbation introduction}, colback=blue!10,breakable]
\# Instructions\\
You are given a text with terms marked between square brackets [ ] and double curly braces \{\{ \}\}. Your goal is to modify the terms marked between square brackets [ ] to make the text incorrect, misleading, or omit critical information.\\

Each change must be semantically credible, contextually plausible, and linguistically natural to a non-expert reader. For example, a gas must be replaced with another gas, or a country must be replaced with another country. Avoid over-generalizing substitutions that dilute meaning (e.g., ``gases'' instead of ``greenhouse gases''), unless vagueness is the intended form of misinformation. Also avoid replacing terms with obvious synonyms. Do not invent non-standard terminology or introduce substitutions that would appear absurd, obviously false, grammatically broken, or conceptually incoherent.\\

You can make small adaptations of nearby words ONLY for grammatical correctness. However, all terms between double curly braces \{\{ \}\} MUST remain identical as the input. Remove square brackets [ ] from changed terms. Retain the original sentence structure and style wherever possible.\\

You are given a free space for planning your strategy. For each replacement word, try to list two to three alternatives and why they are good choices before coming up with a final decision. Output this planning between the marks <thinking></thinking>.\\

Finally, produce the final raw output without any further notes, explanations, or formatting between he marks <output></output>.\\

\# Example\\
\textasciigrave\textasciigrave\textasciigrave\\
The ozone layer protects [the Earth] by absorbing [harmful ultraviolet radiation] from [the Sun]. It is \{\{primarily\}\} found in [the stratosphere], a layer of the atmosphere. Concerns about ozone depletion rose in [the 1980s] after [the discovery] of [a hole] over \{\{Antarctica\}\}.\\
\textasciigrave\textasciigrave\textasciigrave\\
<thinking>\\
1. [the Earth]\\
    * Options: “living organisms,” “the biosphere”\\
    * Chosen: the biosphere — plausible and often used in environmental contexts, but shifts focus away from the planet itself to just living systems, subtly distorting the scope of the ozone layer’s protective effect.\\
2. [harmful ultraviolet radiation]\\
    * Options: “unharmful ultraviolet radiation,” “harmful infrared radiation,” “heat energy”\\
    * Chosen: harmful infrared radiation — sounds technical and solar-related, but it's not what the ozone layer blocks. Misleading but plausible.\\
3. [the Sun]\\
    * Options: “deep space,” “solar flares”\\
    * Chosen: deep space — vague and misleading; implies that source of radiation is a general space phenomenon rather than solar-specific.\\
4. [the stratosphere]\\
    * Options: “mesosphere,” “troposphere,” “ionosphere”\\
    * Chosen: troposphere — the lowest layer, where weather happens, not where ozone is concentrated. Still sounds reasonable to a non-expert.\\
5. [the 1980s]\\
     * Options: “the late 1990s” “the 1970s,” “the early 1990s”\\
     * Chosen: the late 1990s — shifts timeline by a bit, particularly when the problem became of concern for the general public.\\
6. [the discovery]\\
     * Options: “a theory,” “an assumption,” “a hypothesis”\\
     * Chosen: a theory — undermines scientific certainty subtly without being absurd.\\
7. [a hole]\\
     * Options: “an irregularity,” “a gap,” “a reduction”\\
     * Chosen: an irregularity — Very neutral, sounds like a small change rather than a serious issue, minimizing severity.\\
</thinking>\\

<output>The ozone layer protects the biosphere by absorbing harmful infrared radiation from deep space. It is \{\{primarily\}\} found in the troposphere, a layer of the atmosphere. Concerns about ozone depletion rose in the late 1990s after a theory of an irregularity over \{\{Antarctica\}\}.</output>
\end{tcolorbox}

\begin{tcolorbox}[title={LLM-as-a-judge}, colback=blue!10]
\#\#\#Task Description:\\
An instruction (might include an Input inside it), a response to evaluate, a reference answer that gets a score of 5, and a score rubric representing a evaluation criteria are given.\\
1. Write a detailed feedback that assess the quality of the response strictly based on the given score rubric, not evaluating in general.\\
2. After writing a feedback, write a score that is an integer between 1 and 5. You should refer to the score rubric.\\
3. The output format should look as follows: "Feedback: (write a feedback for criteria) [RESULT] (an integer number between 1 and 5)"\\
4. Please do not generate any other opening, closing, and explanations.\\

\#\#\#The instruction to evaluate:\\
\{question\}\\

\#\#\#Response to evaluate:\\
\{response\}\\

\#\#\#Reference Answer (Score 5):\\
\{reference\_answer\}\\

\#\#\#Score Rubrics:
Does the response demonstrate factual correctness by covering all essential points from the reference answer without introducing inaccuracies, omissions, or hallucinated information?\\

- Score 1: The response is mostly factually incorrect or misleading, with many inaccuracies or fabricated information.\\
- Score 2: The response contains multiple factual inaccuracies, significant omissions, or introduces misleading/untrue statements that affect the answer quality.\\
- Score 3: The response is mostly factually correct, but has some inaccuracies, omissions, or unsupported information that weakens it.\\
- Score 4: The response is factually correct and mostly complete, with very minor omissions or imprecisions that do not affect the overall understanding.
- Score 5: The response is entirely factually accurate, fully complete based on the instruction, and does not introduce any inaccuracies, fabrications, or unsupported information.\\

\#\#\#Feedback: 
\end{tcolorbox}

%%================================================================

\newpage

\section{Evaluation Interface} \label{appendix:evaluation-interface}
\begin{figure}[h]
\begin{subfigure}[t]{\linewidth}
\includegraphics[width=\linewidth]{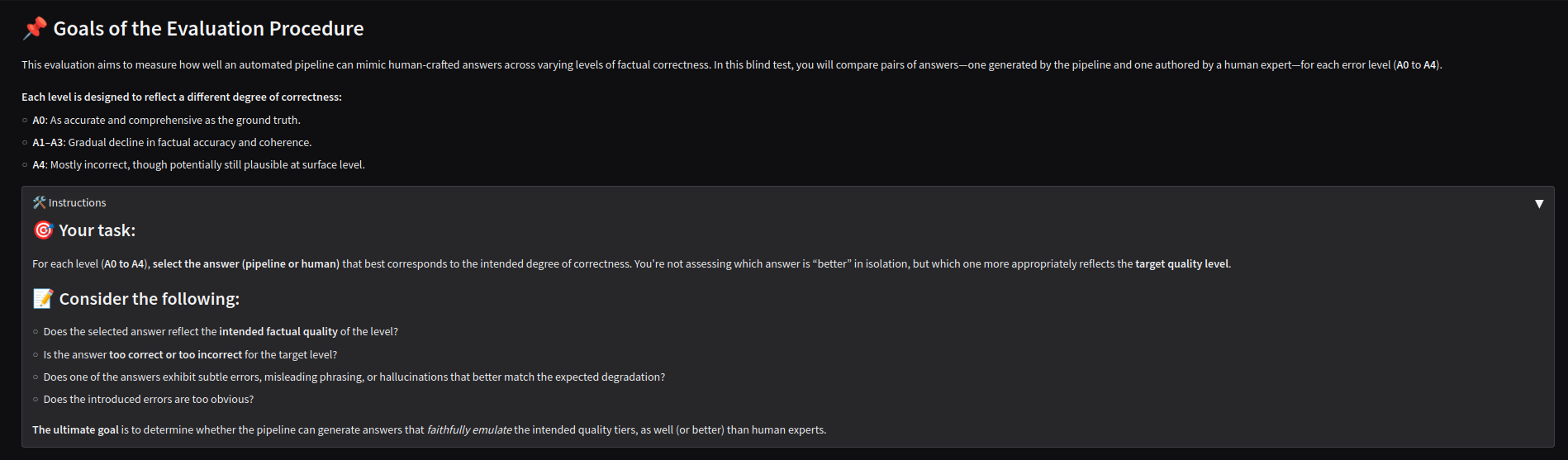}
\caption{Instructions.}
\end{subfigure}
\begin{subfigure}[t]{\linewidth}
\includegraphics[width=\linewidth]{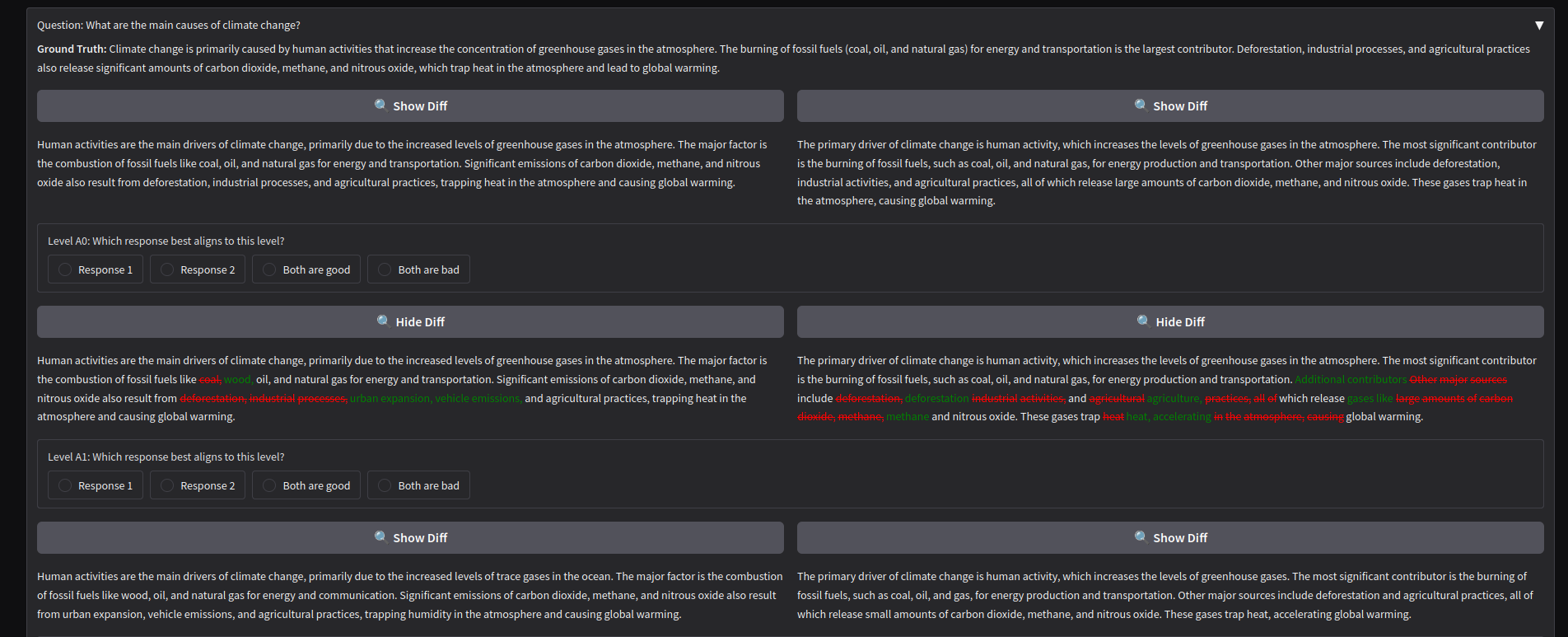}
\caption{Evaluation.}
\end{subfigure}
\caption{Screenshots of the evaluation interface.}
\label{fig:screenshots}
\end{figure}

We ask evaluators to compare the quality of the human expert and the factual perturbation pipeline in a randomized blind test. To facilitate the task, we provide evaluators with an interface that shows scoring guidelines, the questions, and places human and pipeline-introduced perturbations side by side. Screenshots of the interface are available in Figure \ref{fig:screenshots}, and the evaluation guidelines can be found in the box below. For each pair, the evaluator can quickly display the differences between any level and the A0 answer. This feature aims to reduce the cognitive load of tracking changes between levels, enabling the evaluation to focus on the quality of the perturbation.

The evaluators can choose, for each human-pipeline generated pair, which response they think better aligns with the intended level. They are also allowed to reject or accept both if neither or both satisfy the conditions.

\begin{tcolorbox}[title={Evaluation Guidelines}, colback=blue!10]
This evaluation aims to measure how well an automated pipeline can mimic human-crafted answers across varying levels of factual correctness. In this blind test, you will compare pairs of answers — one generated by the pipeline and one authored by a human expert — for each error level (A0 to A4). \\

Each level is designed to reflect a different degree of correctness: \\

    A0: As accurate and comprehensive as the ground truth. \\
    A1–A3: Gradual decline in factual accuracy and coherence. \\
    A4: Mostly incorrect, though potentially still plausible at surface level. \\

\#Your task: \\

For each level (A0 to A4), select the answer (pipeline or human) that best corresponds to the intended degree of correctness. You're not assessing which answer is “better” in isolation, but which one more appropriately reflects the target quality level. \\

\# Consider the following: \\

    Does the selected answer reflect the intended factual quality of the level? \\
    Is the answer too correct or too incorrect for the target level? \\
    Does one of the answers exhibit subtle errors, misleading phrasing, or hallucinations that better match the expected degradation? \\
    Does the introduced errors are too obvious? \\

The ultimate goal is to determine whether the pipeline can generate answers that faithfully emulate the intended quality tiers, as well (or better) than human experts.
\end{tcolorbox}

\clearpage
\newpage

\section{Dataset Examples} \label{appendix:dataset-examples}

\begin{figure}[h]
\centering
\begin{tcolorbox}[colback=gray!5!white,colframe=black!75!white,
  title=Example -- \textit{When did Korea separate into North and South?}, fonttitle=\bfseries,
  boxrule=0.8pt, arc=4pt, top=4pt, bottom=4pt, left=6pt, right=6pt]
\footnotesize
\begin{itemize}
    \item \textbf{A0 (Reference)}: After \textcolor{teal}{World War II}, Korea was divided into North and South Korea, marking \textcolor{teal}{the end} of the Japanese Empire's \textcolor{teal}{35-year occupation} of Korea \textcolor{teal}{in 1945}. The United States and the Soviet Union each \textcolor{teal}{took control} of separate areas of \textcolor{teal}{the country}, with the dividing line at \textcolor{teal}{the 38th parallel}.

    \item \textbf{A1 (Low perturbation)}: After \textcolor{teal}{World War II}, Korea was divided into North and South Korea, marking \textcolor{teal}{the end} of the Japanese Empire's \textcolor{red}{25-year occupation} of Korea \textcolor{teal}{in 1945}. The United States and the Soviet Union each \textcolor{teal}{took control} of separate areas of \textcolor{teal}{the country}, with the dividing line at \textcolor{red}{the 36th parallel}.

    \item \textbf{A2 (Medium perturbation)}: After \textcolor{teal}{World War II}, Korea was divided into North and South Korea, marking \textcolor{red}{a shifting} of the Japanese Empire's \textcolor{red}{25-year occupation} of Korea \textcolor{teal}{in 1945}. The United States and the Soviet Union each \textcolor{teal}{took control} of separate areas of \textcolor{teal}{the country}, with the dividing line at \textcolor{red}{the 36th parallel}.

    \item \textbf{A3 (High perturbation)}: After \textcolor{teal}{World War II}, Korea was divided into North and South Korea, marking \textcolor{red}{a shifting} of the Japanese Empire's \textcolor{red}{25-year occupation} of Korea \textcolor{red}{in 1950}. The United States and the Soviet Union each \textcolor{red}{took influence} of separate areas of \textcolor{teal}{the country}, with the dividing line at \textcolor{red}{the 36th parallel}.

    \item \textbf{A4 (Extreme perturbation)}: After \textcolor{red}{World War I}, Korea was divided into North and South Korea, marking \textcolor{red}{a shifting} of the Japanese Empire's \textcolor{red}{25-year occupation} of Korea \textcolor{red}{in 1950}. The United States and the Soviet Union each \textcolor{red}{took influence} of separate areas of \textcolor{red}{the region}, with the dividing line at \textcolor{red}{the 36th parallel}.
\end{itemize}
\end{tcolorbox}
\end{figure}

\begin{figure}[h]
\centering
\begin{tcolorbox}[colback=gray!5!white,colframe=black!75!white,
  title=Example -- \textit{Who breaks a tie in the US Senate?}, fonttitle=\bfseries,
  boxrule=0.8pt, arc=4pt, top=4pt, bottom=4pt, left=6pt, right=6pt]
\footnotesize
\begin{itemize}
\item \textbf{A0 (Reference)}: As outlined in \textcolor{teal}{Article I}, \textcolor{teal}{Section 3}, Clause 4 of \textcolor{teal}{the United States Constitution}, the Vice President serves as the \textcolor{teal}{ex officio President} of the Senate but is \textcolor{teal}{only} permitted to vote to resolve a tie. As per the U.S. Senate, up to \textcolor{teal}{February 28, 2018}, 36 vice presidents had cast a tie-breaking vote \textcolor{teal}{a total of 264} times.

    \item \textbf{A1 (Low perturbation)}: As outlined in \textcolor{red}{Article II}, \textcolor{teal}{Section 3}, Clause 4 of \textcolor{teal}{the United States Constitution}, the Vice President serves as the \textcolor{teal}{ex officio President} of the Senate but is \textcolor{teal}{only} permitted to vote to resolve a tie. As per the U.S. Senate, up to \textcolor{teal}{February 28, 2018}, 36 vice presidents had cast a tie-breaking vote \textcolor{red}{approximately} \textcolor{teal}{264} times.

    \item \textbf{A2 (Medium perturbation)}: As outlined in \textcolor{red}{Article II}, \textcolor{red}{Part 3}, Clause 4 of \textcolor{teal}{the United States Constitution}, the Vice President serves as the \textcolor{teal}{ex officio President} of the Senate but is \textcolor{teal}{only} permitted to vote to resolve a tie. As per the U.S. Senate, up to \textcolor{teal}{February 28}, \textcolor{red}{2017}, 36 vice presidents had cast a tie-breaking vote \textcolor{red}{approximately 275} times.

    \item \textbf{A3 (High perturbation)}: As outlined in \textcolor{red}{Section II, Part 3}, Clause 4 of \textcolor{teal}{the United States Constitution}, the Vice President serves as an \textcolor{red}{honorary member} of the Senate but is \textcolor{teal}{only} permitted to vote to resolve a tie. As per the U.S. Senate, up to \textcolor{red}{March 28, 2017}, 36 vice presidents had cast a tie-breaking vote \textcolor{red}{approximately 275} times.

    \item \textbf{A4 (Severe perturbation)}: As outlined in \textcolor{red}{Section II, Part 3}, Clause 4 of \textcolor{red}{federal law}, the Vice President serves as an \textcolor{red}{honorary member} of the Senate but is \textcolor{red}{freely} permitted to vote to resolve a tie. As per the U.S. Senate, up to \textcolor{red}{March 25, 2017}, 36 vice presidents had cast a tie-breaking vote \textcolor{red}{approximately 275} times.

\end{itemize}
\end{tcolorbox}
\end{figure}

\begin{figure}[h]
\centering
\begin{tcolorbox}[colback=gray!5!white,colframe=black!75!white,
  title=Example -- \textit{Who wrote the text for ``Jeanie with the Light Brown Hair?''}, fonttitle=\bfseries,
  boxrule=0.8pt, arc=4pt, top=4pt, bottom=4pt, left=6pt, right=6pt]
\footnotesize
\begin{itemize}
    \item \textbf{A0 (Reference)}: ``Jeanie with the Light Brown Hair'' is \textcolor{teal}{a parlor song} created by \textcolor{teal}{Stephen Foster (1826–1864)}, published by \textcolor{teal}{Firth}, \textcolor{teal}{Pond \& Co.} \textcolor{teal}{in New York} \textcolor{teal}{in 1854}. Foster composed the song thinking of his \textcolor{teal}{estranged wife}, \textcolor{teal}{Jane McDowell}, and the lyrics hint at an enduring separation.

    \item \textbf{A1 (Low perturbation)}: ``Jeanie with the Light Brown Hair'' is \textcolor{teal}{a parlor song} created by \textcolor{red}{Henry Bishop (1810–1880)}, published by \textcolor{teal}{Firth, Pond \& Co.} \textcolor{red}{in Philadelphia} \textcolor{teal}{in 1854}. Foster composed the song thinking of his \textcolor{teal}{estranged wife}, \textcolor{teal}{Jane McDowell}, and the lyrics hint at an enduring separation.

    \item \textbf{A2 (Medium perturbation)}: ``Jeanie with the Light Brown Hair'' is \textcolor{teal}{a parlor song} created by \textcolor{red}{Henry Bishop (1810–1880)}, published by \textcolor{teal}{Firth}, \textcolor{red}{Lake \& Co.} \textcolor{red}{in Philadelphia} \textcolor{teal}{in 1854}. Foster composed the song thinking of his \textcolor{teal}{estranged wife}, \textcolor{red}{Anna Brown}, and the lyrics hint at an enduring separation.

    \item \textbf{A3 (High perturbation)}: ``Jeanie with the Light Brown Hair'' is \textcolor{teal}{a parlor song} song created by \textcolor{red}{Henry Bishop (1810–1880)}, published by \textcolor{red}{Morris}, \textcolor{red}{Lake \& Co.} \textcolor{red}{in Philadelphia} \textcolor{teal}{in 1854}. Foster composed the song thinking of an \textcolor{red}{old friend}, \textcolor{red}{Anna Brown}, and the lyrics hint at an enduring separation.

    \item \textbf{A4 (Extreme perturbation)}: ``Jeanie with the Light Brown Hair'' is \textcolor{red}{a folk song} created by \textcolor{red}{Henry Bishop (1810–1880)}, published by \textcolor{red}{Morris}, \textcolor{red}{Lake \& Inc.} \textcolor{red}{in Philadelphia} \textcolor{red}{in 1860}. Foster composed the song thinking of an \textcolor{red}{old friend}, \textcolor{red}{Anna Brown}, and the lyrics hint at an enduring separation.
\end{itemize}
\end{tcolorbox}
\end{figure}

\end{document}